%% file: 0-main.tex
\documentclass{vgtc}                          

\input{_settings}
\input{_macros}

\graphicspath{{figures/}{pictures/}{images/}{./}} 

\usepackage{times}                     

\usepackage{tabu}                      
\usepackage{booktabs}                  
\usepackage{lipsum}                    
\usepackage{mwe}                       
\usepackage[utf8]{inputenc}
\usepackage{amsmath}
\usepackage{amssymb}
\usepackage{algorithm}
\usepackage{algpseudocode}
\usepackage{mathptmx}                  

\onlineid{1430}

\vgtccategory{Research}

\vgtcinsertpkg

\ieeedoi{xx.xxxx/xxxx.20xx.xxxxxxx}

\title{\sysname: Deformation-based Adaptive Volumetric Video Streaming}

\author{Boyan Li%
\and Yongting Chen %
\and Dayou Zhang
\and Fangxin Wang}
\affiliation{\scriptsize The Chinese University of Hong Kong, Shenzhen}

\abstract{
    Volumetric video streaming offers immersive 3D experiences but faces significant challenges due to high bandwidth requirements and latency issues in transmitting detailed content in real time. Traditional methods like point cloud streaming compromise visual quality when zoomed in, and neural rendering techniques are too computationally intensive for real-time use. Though mesh-based streaming stands out by preserving surface detail and connectivity, offering a more refined representation for 3D content, traditional mesh streaming methods typically transmit data on a per-frame basis, failing to take full advantage of temporal redundancies across frames. This results in inefficient bandwidth usage and poor adaptability to fluctuating network conditions. We introduce \sysname, a novel framework that enhances volumetric video streaming performance by leveraging the inherent deformability of mesh-based representations. DeformStream uses embedded deformation to reconstruct subsequent frames from inter-frame motion, significantly reducing bandwidth usage while ensuring visual coherence between frames. To address frame reconstruction overhead and network adaptability, we formulate a new QoE model that accounts for client-side deformation latency and design a dynamic programming algorithm to optimize the trade-off between visual quality and bandwidth consumption under varying network conditions. Our evaluation demonstrates that \sysname outperforms existing mesh-based streaming systems in both bandwidth efficiency and visual quality, offering a robust solution for real-time volumetric video applications.
} 


\begin{document}


\firstsection{Introduction}

\maketitle

\input{1-intro}

\input{2-motivation}
\input{3-design}
\input{4-algorithm}
\input{5-evaluation}

\input{6-related}
\input{7-conclusion}


\bibliographystyle{abbrv-doi}

\bibliography{references,template}
\end{document}

%% file: _settings.tex
\usepackage{pifont}
\usepackage{xspace}
\usepackage{amsmath}
\usepackage{cite}

\newcommand{\sysname}{DeformStream\xspace}

%% file: _macros.tex
\newcommand{\head}[1]{\vspace{0.5mm}\noindent{\bf #1:}}



%% file: 1-intro.tex
Recent advancements in immersive media technologies have significantly expanded the possibilities for volumetric video streaming, which allows viewers to experience three-dimensional scenes from virtually any angle or position. This technology has found applications in a wide range of fields, including live performance streaming, online education, and virtual reality (VR), offering users an unprecedented level of interaction and realism. However, the transmission of volumetric video data over the Internet presents substantial challenges. Chief among these is the enormous bandwidth required to deliver such rich, detailed content in real-time, which has traditionally limited the practicality and scalability of volumetric video streaming in dynamic, real-world environments.

Conventional approaches to volumetric video streaming have primarily relied on point cloud-based techniques~\cite{zhang_yuzu_nodate,han_vivo_2020,lee_groot_2020,liu_vues_2022,lee_farfetchfusion_2023,hu_livevv_2023}. While point clouds offer a relatively lightweight format that facilitates real-time processing and low rendering costs, they require larger amount of points being transmitted for visual coherence~\cite{viola_volumetric_2022}, particularly when zoomed in, due to the lack of surface detail and connectivity between points. This results in a significant trade-off between data efficiency and visual quality, making point cloud-based systems less effective in applications that demand high fidelity and seamless user experiences. The latest neural-network-based approaches, NeRF~\cite{mildenhall_nerf_2020} and 3D Gaussian~\cite{kerbl_3d_2023} splatting generate 2D views by learning the implicit neural representation of 3D scenes, but they require extensive training time for scene reconstruction and rendering~\cite{tosi_how_2024}, making them unrealistic for display on a real-time end-to-end live streaming basis. 

Mesh-based streaming methods that represent surfaces through a network of vertices and faces can provide superior visual quality by maintaining surface detail and connectivity. Meshes offer a structured data format that not only supports detailed surface representation but also facilitates complex deformability, making them highly adaptable for streaming dynamic scenes.
However, current mesh-based 3D sequence compression works originate from the computer vision research community, focusing on the reconstruction quality~\cite{maglo_3d_2015}, while existing mesh-based streaming adaptation techniques consider either intra-frame compression~\cite{crowle_dynamic_2015} or hardware workload~\cite{athanasoulis_optimizing_nodate}. They have several shortcomings: Firstly, they often process and transmit mesh data on a per-frame basis, failing to exploit temporal redundancies between frames, leading to unnecessary bandwidth consumption. Secondly, they failed to consider the mesh's inter-frame deformability to empower visual coherence. Thirdly, existing approaches lack robust network adaptability and may suffer from playback latency issues, compromising the user experience in fluctuating network conditions.

\begin{figure}[h]
    \centering
    \includegraphics[width=\linewidth]{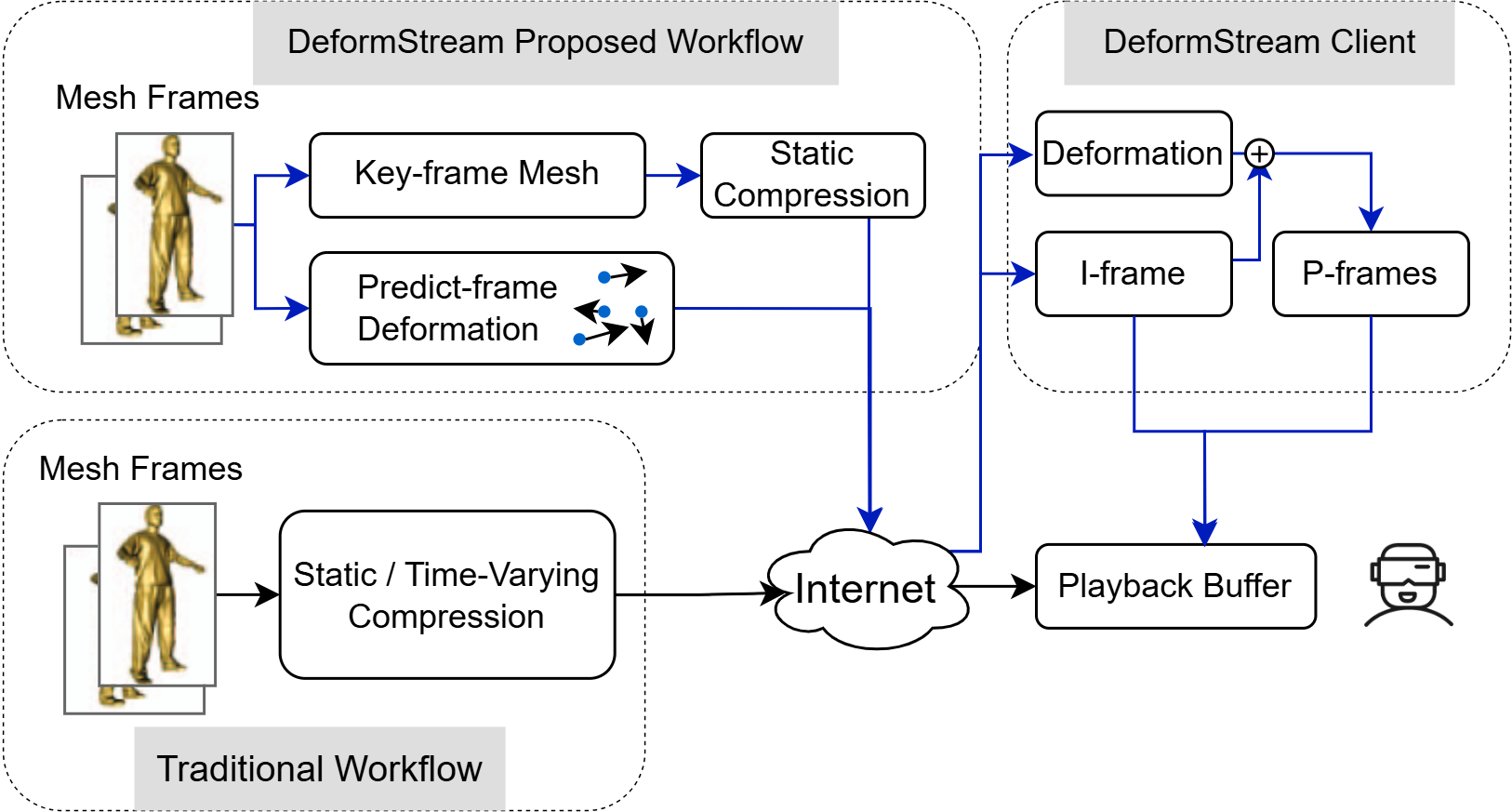}
    \caption{Traditional v.s. proposed streaming pipeline.}
    \label{fig:traditional-vs-proposed}
\end{figure}

In this paper, we introduce \sysname, a novel framework designed to enhance the efficiency and quality of volumetric video streaming by leveraging the inherent visual consistency advantage of the deformability in mesh-based data representations. 
As shown in Figure~\ref{fig:traditional-vs-proposed}, DeformStream utilizes embedded deformation to reconstruct the next frame by such inter-frame motion, significantly improving the streaming of dynamic scenes by ensuring motion coherence between frames. 

\input{2-motivation-comparison-table}

To make our proposed framework practical, we address two key challenges in the streaming pipeline: frame reconstruction overhead and frame transmission adaptation.
To meditate on the latency overhead introduced by reconstruction, we formulate a comprehensive QoE model that includes the buffer refilling time introduced by client-side deformation and formulate a frame adaptation problem to optimize the QoE.
For frame adaptation, we formulate the problem as an optimization problem, where we design a dynamic programming-based algorithm to optimize the trade-off between visual quality and bandwidth usage under varying network conditions.

The technical contributions of this work are threefold:

\begin{itemize}
    \item We propose a deformation-based approach to implement inter-frame compression in volumetric streaming systems. We propose a deformation-based adaptive streaming method that dynamically adjusts to network conditions by optimizing the balance between visual quality and bandwidth consumption.
    \item We develop a dynamic programming algorithm to effectively solve the frame adaptation problem, ensuring smooth and uninterrupted playback even under challenging network environments.
    \item Our evaluation demonstrates that \sysname outperforms existing mesh-based streaming systems in both bandwidth efficiency and visual quality, offering a more robust solution for real-time volumetric video applications.
\end{itemize}





%% file: 2-motivation-comparison-table.tex
\begin{table*}[]
    \centering
    \begin{tabular}{c|c|c|c|c|c}
        \toprule
        3D Representation  & Processing/Training Cost & Rendering Cost & Transmission Size & Deformability/Editing & Visual Quality  \\ \midrule
        Point Cloud & Real-Time~\cite{kowalski_livescan3d_2015, inproceedings} & Low & Medium & Medium & Medium  \\ \hline
        Mesh & Real-Time~\cite{jin_meshreduce_2024} & Low & Medium & High & High  \\ \hline
        NeRF & Slow~\cite{mildenhall_nerf_2020,tosi_how_2024} & High & High & Low & High  \\ \hline
        3D Gaussian Splatting & Slow~\cite{kerbl_3d_2023} & Medium & High & Low & High \\ \bottomrule
    \end{tabular}
    \label{tab:comparison-3d-repr}
    \caption{Comparison of 3D Representations in Streaming Scenarios}
\end{table*}

%% file: 2-motivation.tex
\section{Background and Motivation}


In this section, we first discuss the advantages of using mesh as the 3D representation in streaming and the deficiencies of current mesh-based streaming systems. We then discuss existing incremental video streaming approaches for time-varying 3D content and mesh-based volumetric video streaming systems. Finally, we summarize the limitations of existing works and motivate our deformation-based approach. 

\subsection{3D Representations for Streaming}

Recent research has explored various representations for volumetric video streaming, each with distinct data formats that cater to different needs. We compare these 3D representations in the context of streaming and playback. 

As depicted in Table~\ref{tab:comparison-3d-repr}, point clouds are composed of discrete points $(x,y,z)$ and $(r,g,b)$ which offer a lightweight format that enables real-time processing and low rendering costs, but they often result in poor visual quality when zooming in due to the lack of surface detail and connectivity between points. Meshes, which represent surfaces through a network of vertices, edges, and faces, provide a more structured data format that supports detailed surface representation and allows for high visual quality. This structure also facilitates complex deformability and editing, making meshes highly adaptable for dynamic scenes, allowing one to reconstruct new mesh from simple keypoint correspondences~\cite{sumner_embedded_2007}. Neural Radiance Fields (NeRFs) utilize a neural network to encode volumetric data, achieving high visual quality but at the cost of slow processing and high rendering requirements, as they require intensive computation to generate views from different angles~\cite{mildenhall_nerf_2020,tosi_how_2024}. Similarly, 3D Gaussian representations, which model scenes using Gaussian distributions, offer smooth transitions and high visual fidelity but involve much slower training and moderate rendering costs~\cite{kerbl_3d_2023,tosi_how_2024}. 

Given the need for a balance between high visual quality, real-time processing, and efficient transmission, we focus on the mesh format in this work. Meshes offer a structured yet flexible data format that supports both \textit{high-quality rendering} and \textit{efficient processing}, which is ideal for real-time 3D video streaming applications.


\subsection{Incremental Video Streaming}

In 3D video streaming, the adjacent frame differences are typically minimal, and significant redundancy exists in the per-frame approach. There are recent advances in 3D incremental video streaming to leverage spatial similarity and minimize unnecessary transmission costs. We categorize them into two types: 

\head{(1) Codec-level} A benchmark has compared several open-source compression schemes according to their compression rate and distortion, and turns out that Draco~\cite{google_googledraco_2024} and O3DGC~\cite{sheth_rbshethopen3dgc_2024} have the best-performing rate-distortion trade-off~\cite{doumanoglou_benchmarking_2019}. However, the impact of reconstruction time—an important factor in user experience—is still neglected. This limitation motivates the need for a deformation-based approach that efficiently balances compression with streaming adaptability, ensuring a seamless user experience.

\head{(2) Block-level content reusing} Due to the representation sparsity, point cloud-based volumetric video streaming systems Hermes~\cite{wang_hermes_2023} and CaV3~\cite{liu_cav3_2023} make use of temporal similar cells to reduce re-transmission of those cells poses high-similarity. However, this approach cannot be applied to mesh data structure since the vertex connectivity cannot be broken apart into blocks.




\subsection{Mesh Video Streaming}
Previous works have explored how mesh sequences are reconstructed and transmitted from capture to streaming.
Holoportation~\cite{orts-escolano_holoportation_2016} and SLAMCast~\cite{stotko_slamcast_2019} captures the scene using multiple RGB-Depth cameras and streams the reconstructed objects within the local network.
Live4D~\cite{zhou_live4d_2023} and MeshReduce~\cite{jin_meshreduce_2024} are the first two real-time volumetric video capture and streaming systems that use mesh as their representation, both of them use a per-frame RGBD-to-mesh reconstruction pipeline approach that streams per-frame compressed mesh sequence to the client. 

Previous research on mesh-based network adaptation has focused on finding efficient coding solutions to balance bandwidth requirements and low-latency streaming. For instance, one approach involved designing a network monitor-compression parameter tuning adaptation service to reduce data requirements~\cite{crowle_dynamic_2015}. This method monitored network metrics such as packet loss, bit and frame rates in both sender and receiver, as well as user-perceived Quality of Experience (QoE) (Athanasoulis). Another approach used a cognitive network optimizer based on reinforcement learning to adjust mesh compression levels based on QoE. Additionally, a serverless framework was designed with a network optimization strategy to balance trade-offs between user-side QoE and content-provider-side transcode cost~\cite{konstantoudakis_serverless_2022}.

\begin{figure}[t]
    \centering
    \includegraphics[width=1\linewidth]{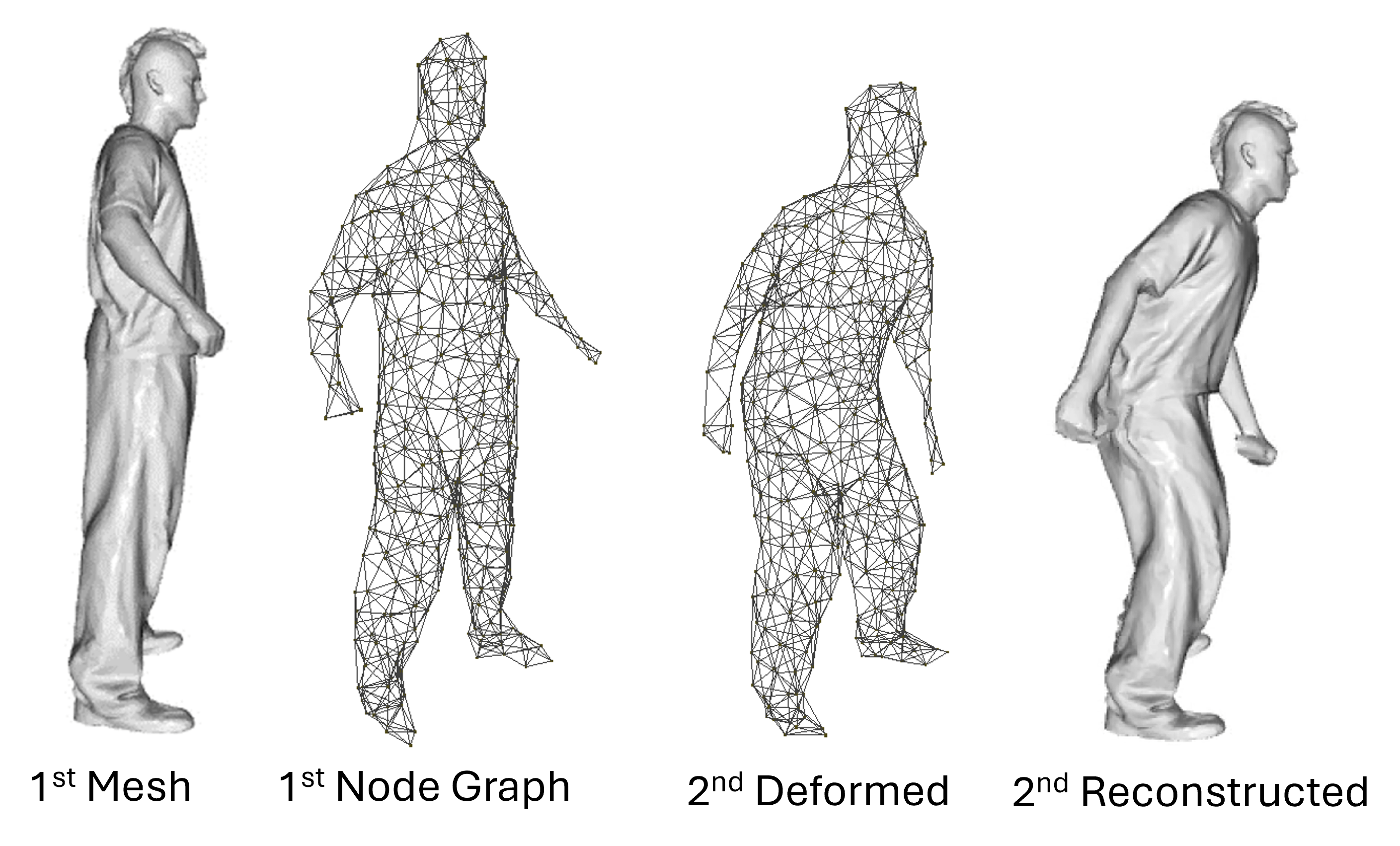}
    \caption{Embedded Deformation Graph}
    \label{fig:emb-def-graph}
\end{figure}

\begin{figure*}[ht]
    \centering
    \includegraphics[width=1\linewidth]{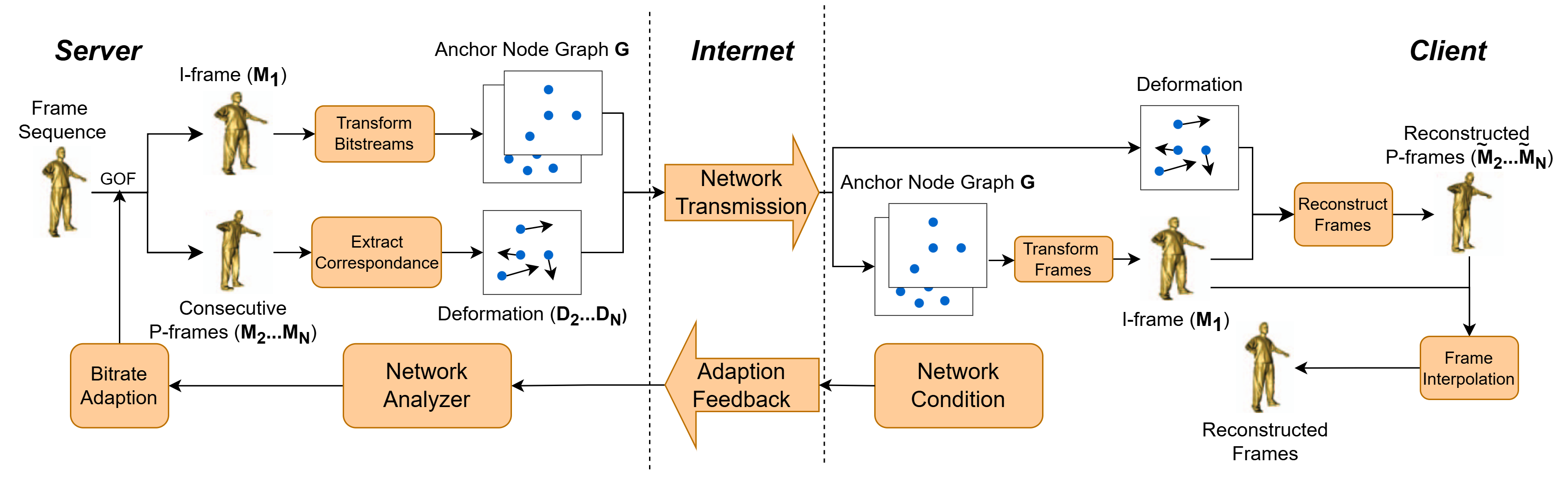}
    \caption{Workflow Overview}
    \label{fig:workflow}
\end{figure*}

However, the deformation ability that mesh offers has been under-explored in these works, which provides both higher smoothness~\cite{sumner_embedded_2007} and preserves inter-frame correlation to reduce frame difference transmission data. As demonstrated in Figure~\ref{fig:emb-def-graph}, the core idea behind the deformation models~\cite{sederberg_free-form_1986,le_interactive_2017} is to use fewer key points to control the displacement of more points, with each key point having different influential weights to the original points~\cite{sumner_embedded_2007}. 

To conclude, though deformation-aware mesh streaming provides quality first and bandwidth-efficient opportunities, enabling deformation-based streaming and real-time playback needs joint considerations of the network condition and the user-perceived QoE.




%% file: 3-design.tex
\section{Design and Analysis}

 

\subsection{Workflow Overview}

The DeformStream workflow focuses on adaptive streaming mechanisms to optimize transmission over varying network conditions. We illustrate the workflow in Figure~\ref{fig:workflow}, where it is divided into three main components: Encoder, Network Adaptation, and Decoder.

\head{Encoding and Decoding}
We stream the whole mesh sequence in a chunk-based approach, and we use the concept of Group of Frames (GoF) in 2D videos to demonstrate our workflow since we split the transmission data on a per GoF basis: (1) Each \textbf{I-frame} contains the ground truth mesh geometry data of the first mesh $M_1$ and the anchor node graph $G$, (2) and the consecutive \textbf{P-frames} only contains the deformation matrix $\{D_2, D_3, \cdots, D_n\}$ of each node. The deformation information is constructed by a correspondence extractor~\cite{yao_fast_2023} which consists of the rotation matrix $\mathbf{R}=\{R_i\in\mathbb{R}^3\} \in \mathbb{R}^{N\times3}$ and the translation matrix $\mathbf{T}=\{T_i\in\mathbb{T}^3\} \in \mathbb{T}^{N\times3}$ for each anchor node $j\in[1,N]$. 
Once the client has received an I-frame or a P-frame within a GoF, the new frame $M_t$ at time $t$ is reconstructed through a deformation operator $M'_t=Deform(M_{t-1}, \mathbf{R}, \mathbf{T}, \mathbf{n})$. We present more detail in preliminaries \S\ref{sec:preliminaries-emb-def}, including how to obtain the deformation $\mathbf{R}$ and $\mathbf{T}$ given $M_t$ and $M_{t-1}$.

\begin{figure}[h]
    \centering
    \includegraphics[width=1\linewidth]{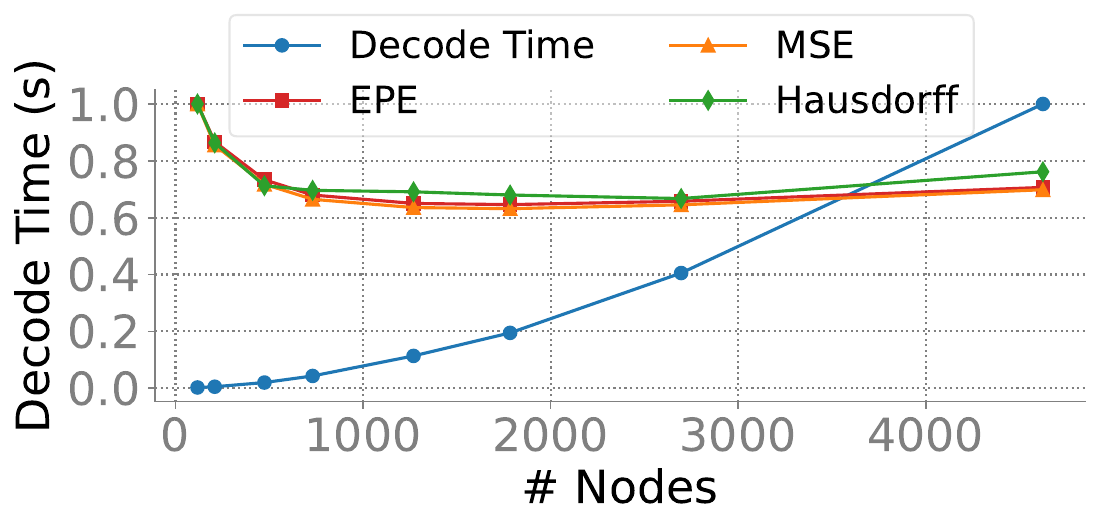}
    \caption{Normalized Decoding Time and Quality Metrics at Different Bitrate}
    \label{fig:node-vs-decoding-time}
\end{figure}

\head{Playback Buffer}
To further adapt the transmission bandwidth under a limited network environment, we set different levels of node graph density $B_{t,b}$ as a variable to control, where $b\in\{L1,L2,L3,\cdots\}$ is the selected bitrate for the frame at time $t$. We observe that reconstructing a new frame using different levels of nodes would diverse in the decoding time which slows down the re-buffering time at the client side and it is unknown how the relationship is, we further run deformation tests for $nodes(b)$ from 120 to 4600 to profile the decoding time. As shown in Figure~\ref{fig:node-vs-decoding-time}, the decoding time increases nearly linearly. Another observation is that the geometry error does not continuously lower as the number of nodes grows. We use these two observations to fit the rebuffering time and set the bitrate boundary in the QoE model (\S\ref{sec:qoe-model}).

\head{QoE-aware Network Adaptation}
We use a QoE-aware network adaptation module to identify the elements in the process that could impact the delivery of mesh frames to the end user. The network adaptation problem tries to answer: what are the coding parameters for both texture and geometry to produce optimal quality for a given bitrate? We make decisions in a chunk-level approach and decide whether to use I-frame and which level of bitrate to use for P-frames at a given network available bandwidth and QoE. We present detailed formulation in \S\ref{sec:frame-dependency} and \S\ref{sec:abr}.


\subsection{Embedded Deformation}
\label{sec:preliminaries-emb-def}
In this section, we introduce our encoding and decoding workflow.
The core of embedded deformation lies in its ability to deform space naturally and intuitively by manipulating objects embedded within that space while preserving their local features. This is achieved by constructing a \textit{deformation graph} $G$, where each anchor node in the graph is associated with an affine transformation that controls the deformation of nearby space. The transformations are smoothly blended across the nodes to ensure a globally consistent deformation. To obtain a node graph, nodes are typically sampled from the mesh, with each node connected to others within its influence via undirected edges, forming a network that reflects the spatial relationships between different transformations.

\subsubsection{Encoding}
In each GoF, the I-frame is directly processed and transmitted. Encoding P-frames consists of three steps: node graph extraction, deformation calculation, and deformation graph generation.

\textbf{Embedded node graph extraction:} a sparse set of control nodes is selected from the source shape to represent local deformations while reducing the degrees of freedom. Control nodes are sampled using PCA-based sorting. PCA is applied to the vertex set to determine the major axis along which points are projected and sorted to ensure representative coverage of the source shape. For each vertex \(v_i\), a set of neighboring control nodes is determined based on a fixed influence radius \(R\), forming the set of influencing nodes
\[I(v_i)=\{p_j\in VG | D(v_i, p_j) < R\}\]
where \(D(v_i, p_j)\) represents the geometric distance between vertex \(v_i\) and control node \(p_j\).
Furthermore, edges between control nodes are introduced if they share influence over a common vertex. Specifically, if two nodes both influence vertex \(v_i\), an edge is added between them, forming the edge set
\[EG=\{(p_j, p_k)|p_j, p_k \in I(v_i)\}\]
By combining vertex graph \(I(v_i)\) and edge graph \(EG\), we get the sparse node graph that provides the structural foundation for deformation.

\textbf{Node graph deformation calculation:} Each control node is associated with an affine transformation consisting of rotation, scaling, and translation.
The goal of deformation calculation is to optimize the affine transformations such that the overall deformation aligns the source shape to the target while maintaining smoothness and regularity in the deformations. The total deformation energy consists of three terms: (1) The alignment term ensures that the source shape deforms in a way that minimizes the distance between the deformed source vertices and the target vertices. 
(2) The rotation term encourages the local affine transformations to be as close as possible to rigid transformations (i.e., pure rotations without scaling or shear), 
and (3) the regularization term ensures that neighboring control nodes deform similarly, promoting smooth deformations across the source shape. 
The full optimization problem is
\[\min_{R_{i}, t_{p}} E(X)=\lambda_{Align}E_{align}+\lambda_{Rot}E_{Rot}+\lambda_{Reg}E_{Reg} \forall p_i\in VG\]
The optimization process adjusts the affine matrices \(R_{i}\) and translation vectors \(t_i\) for each control node \(p_i\) in such a way that the deformed source shape closely matches the target shape while maintaining smooth and physically plausible deformations. Each term plays a critical role in ensuring the final deformation is accurate, smooth, and retains the local structural integrity of the shape.

After computing the optimal affine transformations \(R_j\) and translations \(t_j\) for each control node through the optimization process, the final \textbf{deformation graph is generated}. The deformation graph describes the non-rigid deformation of the entire source shape. The deformation of each vertex is influenced by the affine transformations of its neighboring control nodes, ensuring that the overall shape is smoothly deformed towards the target.

\subsubsection{Decoding}
For frames in a GoF, we directly render the received I-frame. For P-frames, we use the node graph and the deformation graph to decode the P-frame. The decoding process is to reconstruct all of the vertices in P frames by deformation of each node in the node graph. The decoding process involves reconstructing all vertices across the P frames by applying the deformation of each node in the node graph. This process relies on a combination of nodes and affine transformations to apply localized deformations to specific regions of a shape. Each node \(j\) extracted by the above Encoding process is associated with a \(3 \times 3\) affine transformation matrix \(R_j\) and a \(3 \times 1\) translation vector \(t_j\), allowing localized manipulation of the surrounding space. The deformation of any vertex \(v_i\) in the vicinity of node \(j\) is expressed as:
\[\tilde{v}_i = R_j(v_i-p_j) + p_j + t_j\]
where \(p_j\) is the position of node \(j\) in space, \(\tilde{v}_i\) is the deformation of vertex \(v_i\), \(t_j \) is the translation vectors for nodes \(j\), representing the translation part of the affine transformation for each node. \(R_j\) is the affine transformation associated with node \(j\). This matrix applies both rotation and scaling transformations to the points near node \(j\).\\
For deforming a geometric model, the final position of each vertex \(v_i\) is computed as a weighted sum of the affine transformations applied by nearby nodes. The deformed position \(\tilde{v}_i\) is given by:
\[\tilde{v}_i=\sum^m_{j=1}w_j(v_i)[R_j(v_i-p_j) + p_j + t_j]\]
where \(w_j\) is the weight that defines the influence of node \(j\) on vertex \(v_i\), in our work, to simplify calculation, all \(w_j\) is set to 1.\\
In addition to transforming the vertices, the deformation graph also handles the transformation of normals to maintain the correct orientation of surface features. The deformed normal \(\tilde{n_i}\) at a vertex \(v_i\) is computed by applying the inverse transpose of the affine transformations associated with the neighboring nodes. The formula for transforming the normal is:
\[\tilde{n}_i=\sum^m_{j=1}w_j(v_i)R_j^{-T}n_i\]
where \(n_i\) is the original normal vector at vertex \(v_i\), \(w_j\) is the weight of node \(j\), and to simplify calculation, we set all \(w_j\) as 1 .\\
By applying the deformation outlined above, the mesh in P-frames is reconstructed by Client.

\subsection{Frame Dependency}
\label{sec:frame-dependency}
To effectively describe the dependency relationships between video frames, we propose a binary segment tree model that hierarchically structures these dependencies. Let the video frame set be defined as \(F=\{f_1, f_2, ..., f_n \} \), where the dependency relationship \(D(f_i, f_j)\) denotes the degree to which frame \(f_j\) depends on frame \(f_i\) , along with the error that arises due to this dependency. We use the binary segment tree to model this hierarchical structure of dependencies, where each node in the tree not only represents the dependency between frames but also quantifies the trade-off between bandwidth and error.

In this tree, the root node represents the entire video frame set FF, where all frames are assumed to depend on the first frame \(f_i\) , a common assumption in many video encoding standards when handling keyframes. Below the root node, each leaf node represents a single frame \(f_i\) , indicating that there is no dependency between frames, implying that each frame is independently encoded. Interval nodes \(\{f_i, f_j\}\) represent a dependency relationship over the interval from frame \(f_i\) to frame \(f_j\) , covering all dependencies from \(D(f_{i-1}, f_i)\) to \(D(f_{j-1}, f_j)\) . This structure enables the tree to capture both global and local dependencies among frames.

One of the key advantages of this model is its ability to flexibly adjust the depth of dependencies between frames, allowing it to adapt to varying network bandwidth conditions. Deep dependencies, represented by the deeper nodes in the tree, indicate strong dependencies between frames, which significantly reduces the error between frames. However, deep dependencies also require higher bandwidth to transmit the necessary information. Therefore, deep dependencies are better suited to high-bandwidth environments, where they can maintain high video quality. In contrast, shallow dependencies, represented by the shallower nodes, denote weaker dependencies, which result in higher error but also lower bandwidth requirements. This makes shallow dependencies more appropriate for bandwidth-constrained scenarios, such as mobile networks or unstable network environments.

This binary segment tree model allows for dynamic adjustment of frame dependencies during video transmission, optimizing the efficiency of adaptive bitrate (ABR) streaming. Specifically, when bandwidth is sufficient, the system can leverage deeper dependency nodes to ensure high video quality. Conversely, when bandwidth is limited, the system can prioritize shallow dependencies to reduce transmission overhead, ensuring smooth and stable playback. 




%% file: 4-algorithm.tex
\section{Deformation-aware Adaptive Network Optimization}
\label{sec:abr}

\subsection{Chunk Level Transmission}

When designing a volumetric video with a frame rate (FPS) of 30, each data chunk contains 30 frames of video content, and the system transmits one chunk per second. This means that each chunk encapsulates one full second of video data, with the transmission rate synchronized to a frequency of 1 chunk per second. As a result, the network must handle a constant bitrate, where each chunk is delivered in real-time to maintain smooth 30 FPS playback. However, encapsulating multiple frames within a single chunk introduces challenges related to error handling, chunk loss, and latency, as the loss of a single chunk would lead to the loss of an entire second of video content.

\subsection{QoE Model}
\label{sec:qoe-model}
At each chunk transmitted, the server selects an appropriate bitrate for the upcoming video chunk. The decision is informed by estimates of available bandwidth, Quality of Experience, and network conditions. The QoE model of frame \(i\) described by:
\[QoE=-\mu_1 q_i - \mu_2 |q_i - q_{i-1}| + \mu_3 l_i\]
provides a multi-faceted evaluation of the user experience, where \(q_i\) represents the quality of frame \(i\) , \(|q_i-q_{i-1}|\) captures the impact of quality fluctuations between consecutive frames, and \(l_i\) represents the playback delay associated with frame \(i\) . This model enables the ABR system to take into account not only the quality of the current segment but also the smoothness of quality transitions and playback delay.

The adaptive streaming problem can be formulated as the following optimization problem:
\[max \sum^n_{i=1}QoE_i\]
subject to the constraint:
\[B_i + \sum^j_{k=i+1}B_{T_{kj}} \leq B_t\]
This constraint ensures that the total bandwidth required for the selected quality levels does not exceed the available network bandwidth \(B_t\), thereby preventing rebuffering events or interruptions in video playback.


\subsection{Adaptive Algorithm}

We use dynamic programming to solve this ABR problem. The process of dynamic programming is as follows:

\textbf{State Initialization:} Define the DP array with three dimension \(dp[i][j][k]\), which means the maximum \(QoE\) in the first \(i\) frames when the total bandwidth consumption \(j\) and the reconstruction state \(k\). 
\[0 \leq i \leq n, 0 \leq j \leq S, k \in \{0, 1\} \]
where \(n\) is the number of frames, and \(S\) is total bandwidth.

Initialize the DP array as \(dp[0][j] = -\infty, \forall j\), which means there is no frame that can reconstruct the bandwidth consumption \(j\).

\(dp[0][0] = 0\), which means when there is no frame, the bandwidth consumption is 0, and the total \(QoE\) is 0.\\
\textbf{State Transition:} For each frame \(i\) , we have 3 choices:

Frame \(i\) is not reconstructed by any other frames, and we transmit the original frame \(i\), and add it to the chunk directly. At this time, the state transition is as follows:
\[dp[i][j][0] = max(dp[i][j][0], dp[i-1][j-s[i][k]]+QoE(i)) \]
Limited by bandwidth, frame \(i\) is reconstructed by \(i-1\) frame. The algorithm calculates the total \(QoE\) based on the size and \(QoE\)of the reconstructed frame. Assume the size and \(QoE\) of the reconstructed frame \(i\) is \(s[i']\) and \(QoE[i']\) The state transition is as follows :
\[dp[i][j][1] = max(dp[i][j][1], dp[i-1][j-s[i]'[k]]+QoE(i)') \]
\textbf{Termination:} After processing all frames, we extract the optimal solution from the dynamic programming array \(dp[n][j][k] \) where \(n\) is the total number of frames. The goal is to find the maximum \(QoE\) while satisfying the total bandwidth constraint SS. Specifically, we select the state with the highest \(QoE\) across all possible bandwidth consumption \(j\) and reconstruction states \(k \{0,1\}\)

In the termination process, the algorithm first iterates through \(dp[n][j][k] \) to find the maximum \(QoE\) such that \(j \leq S\). Once the state with the maximum \(QoE\)  is identified, backtrack through the dynamic programming array to reconstruct the sequence of decisions (i.e., whether each frame was transmitted directly or reconstructed from the previous frame).

Let the final maximum \(QoE\)  be \(QoE_{max}\) , the corresponding bandwidth consumption be \(j_{opt}\) , and the reconstruction state be \(k_{opt}\) . The optimal solution is determined by:

Starting from the state \(dp[n][j_{opt}][k_{opt}] \), backtrack along the stored optimal path to determine the reconstruction strategy for each frame.

Finally, output the maximum \(QoE\) and the corresponding transmission strategy for the frames. 
\begin{figure*}[t]
    \centering
    \begin{minipage}[b]{0.24\textwidth}
    \centering
    \includegraphics[height=3cm]{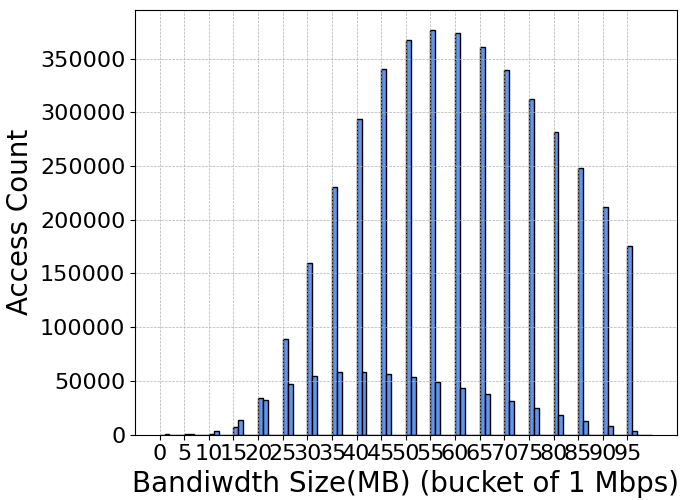}
    \caption{Access Time in each Bandwidth}
    \label{access time in each bandwidth}
    \end{minipage}
    \hfill
    \centering
    \begin{minipage}[b]{0.25\textwidth}
        \centering
        \includegraphics[height=3cm]{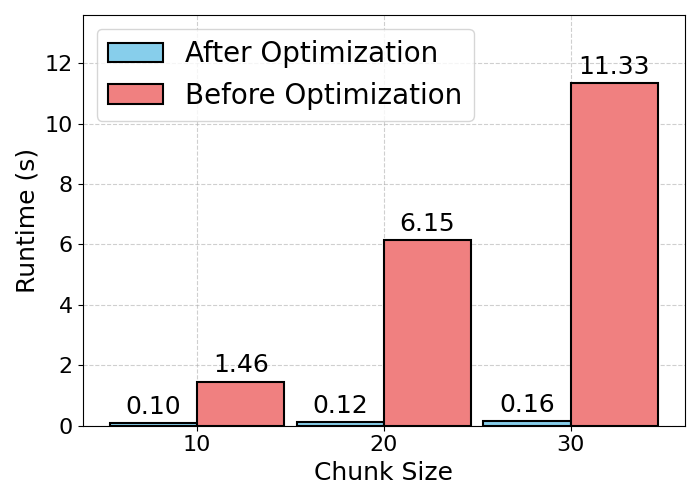}
        \caption{Adaption Algorithm Runtime}
        \label{Performance of Adaption Algorithm}
    \end{minipage}
    \hfill
    \centering
    \begin{minipage}[b]{0.25\textwidth}
        \centering
        \includegraphics[height=3cm]{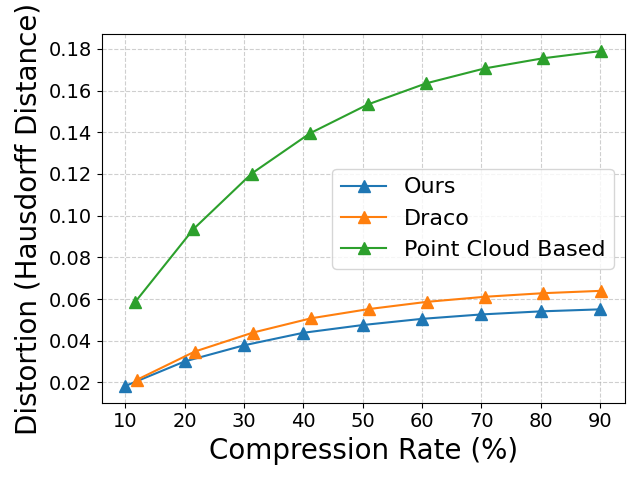}
        \caption{Rate Distortion at Varying Compression Rate}
        \label{Rate Distortion Curve}
    \end{minipage}
    \hfill
    \centering
    \begin{minipage}[b]{0.24\textwidth}
        \includegraphics[height=3cm]{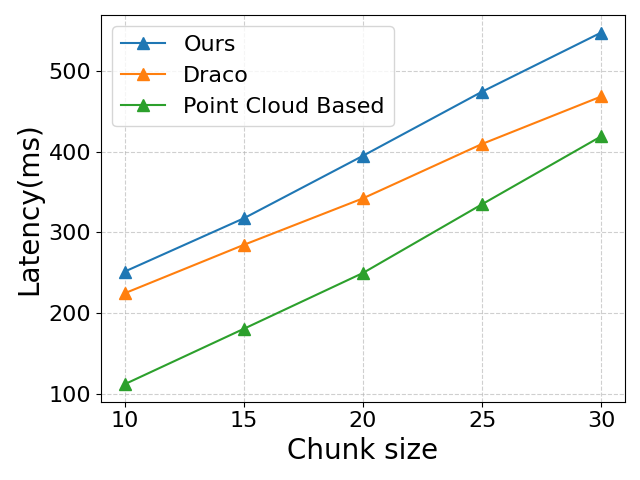}
        \caption{Latency Performance under Different Chunk Size}
        \label{Latency Performance}
    \end{minipage}
\end{figure*}

\subsection{Performance Optimization}

\head{Pruning} We use bounded pruning to improve the efficiency of solving the above adaptive problem by using the bandwidth limitation \(S\). It works by eliminating any states or sequences of frames where the total size exceeds a predefined limit, \(S\). If the bandwidth of a particular frame sequence is greater than \(S\), further exploration of that branch is unnecessary, and the branch is pruned.

This pruning technique reduces computational complexity by restricting the search space, allowing the algorithm to focus only on valid solutions. As a result, it accelerates the process of finding optimal or near-optimal solutions without compromising accuracy, particularly in problems where size or capacity constraints are a critical factor.

\textbf{Sparse Bandwidth Consideration:} From Figure \ref{access time in each bandwidth}, we find that bandwidth is not uniformly distributed. This results in periods where bandwidth availability is extremely limited or fluctuates unpredictably. In such cases, treating bandwidth as a continuous or dense resource becomes inefficient both in modeling and in algorithmic design.

We refer to this situation as sparse bandwidth, where the usable bandwidth SS is only available in small, intermittent quantities. For instance, only a limited set of bandwidth values can be utilized, and the majority of potential bandwidth values remain unused. This sparsity arises due to network bottlenecks, congestion, or prioritized allocation to other services.

Traditional dynamic programming approaches assume that bandwidth can take any value up to SS, leading to an unnecessarily large DP table that allocates resources for states that will never be visited. This results in significant memory overhead and redundant computations.

To address this inefficiency, we model the bandwidth as a sparse resource and design our algorithm to dynamically allocate states only when the corresponding bandwidth values are available. This allows us to effectively optimize the usage of the sparse bandwidth, significantly reducing both the memory footprint and computational complexity of our solution.

\textbf{Multiprocessing Optimization:} Since each frame's choice (compress or not) is relatively independent, and the state transition of each frame depends only on the state of the previous frame, the computation tasks for each frame can be parallelized, significantly accelerating the overall process. Specifically, in this work, multiprocessing was employed to partition the compression state decisions for the 30 frames into multiple subtasks, which are then distributed across several CPU cores for parallel execution. Each process independently computes the compression states and bandwidth usage for its assigned frames, which substantially reduces the computation time.

We provide the pseudo-code of our dynamic programming algorithm in the adaptive system as algorithm \ref{ABR}.

\begin{algorithm}[h]
\caption{Maximize Mesh Quality}
\begin{algorithmic}[1em]
\State \textbf{Input:} $n$, $S$, sizes, src\_meshes, dst\_meshes, combined\_sizes, combined\_rmses
\State $dp \gets \{(0, 0, 0) \mapsto 0\}$ \Comment{Initialize DP table}
\State $prev \gets \{\}$ \Comment{Store path}
\State $min\_rmse \gets \infty$
\For{$i \gets 1$ \textbf{to} $n$}
    \State $next\_dp \gets \{\}$
    \ForAll{$(i\_prev, current\_size, k), current\_rmse \in dp$}
        \State Complete following tasks by multiprocessing
        \If{$i\_prev = i - 1$}
            \State $new\_size \gets current\_size + sizes[i-1]$
            \If{$new\_size \leq S$}
                \State $new\_state \gets (i, new\_size, 0)$
                \State $new\_QoE \gets current\_QoE$
                \If{$new\_QoE \geq next\_dp[new\_state] $}
                \State $next\_dp[new\_state] \gets new\_QoE$
                \State $prev \gets (i\_prev, current\_size, k)$
                \EndIf
            \EndIf
            \If{$current\_size + combined\_sizes[i-1] \leq S$}
                \State Reconstruct Mesh
                \State Compute $new\_QoE$ and $new\_size$
                \State $new\_state \gets (i, new\_size, 1)$
                \If{$new\_QoE \geq next\_dp[new\_state] $}
                \State $next\_dp[new\_state] \gets new\_QoE$
                \State $prev \gets (i\_prev, current\_size, k)$
                \EndIf
            \EndIf
        \EndIf
    \EndFor
    \State $dp \gets next\_dp$
\EndFor
\State Find minimal RMSE from $dp$ and reconstruct the path from prev
\State \textbf{Output:} $min\_rmse$, $path$
\end{algorithmic}
\label{ABR}
\end{algorithm}

\subsection{Performance Comparison}

Figure \ref{Performance of Adaption Algorithm} illustrates the comparison of our Adaption Algorithm runtime before and after performance optimization for different chunk sizes ($size\in\{10, 20, 30\}$). It is evident that the runtime before optimization increases significantly with larger chunk sizes, reaching 11.33 seconds for a chunk size of 30. In contrast, after optimization, the runtime remains almost constant across all chunk sizes, with a maximum of just 0.12 seconds. This demonstrates that the optimization greatly reduces the runtime, particularly for larger data chunks, where the performance improvement is most pronounced. Specifically, for a chunk size of 30 (i.e. 30FPS), the optimized algorithm reduces the runtime by over \(99\%\) , from 11.33 seconds to 0.12 seconds, indicating a substantial increase in both efficiency and scalability.



%% file: 5-evaluation.tex
\section{Evaluation}

\subsection{Experiment Implementation and Setup}
We adopt a modified version of AMM-NRR~\cite{yao_fast_2023} for node graph correspondence extraction and implement the adaptation algorithm in Python and C++. We use a server with an Intel Xeon CPU @2.3GHz and 64GiB memory, and a client with an Intel Core CPU @3.9GHz and 16GiB memory. All experiments in our approach are running single-threaded and reach a 30FPS real-time playback performance. The transmitted original mesh is compressed with Draco~\cite{google_googledraco_2024} for comparison with other approaches.
\begin{figure*}[t]
    \centering
    \includegraphics[width=1\linewidth]{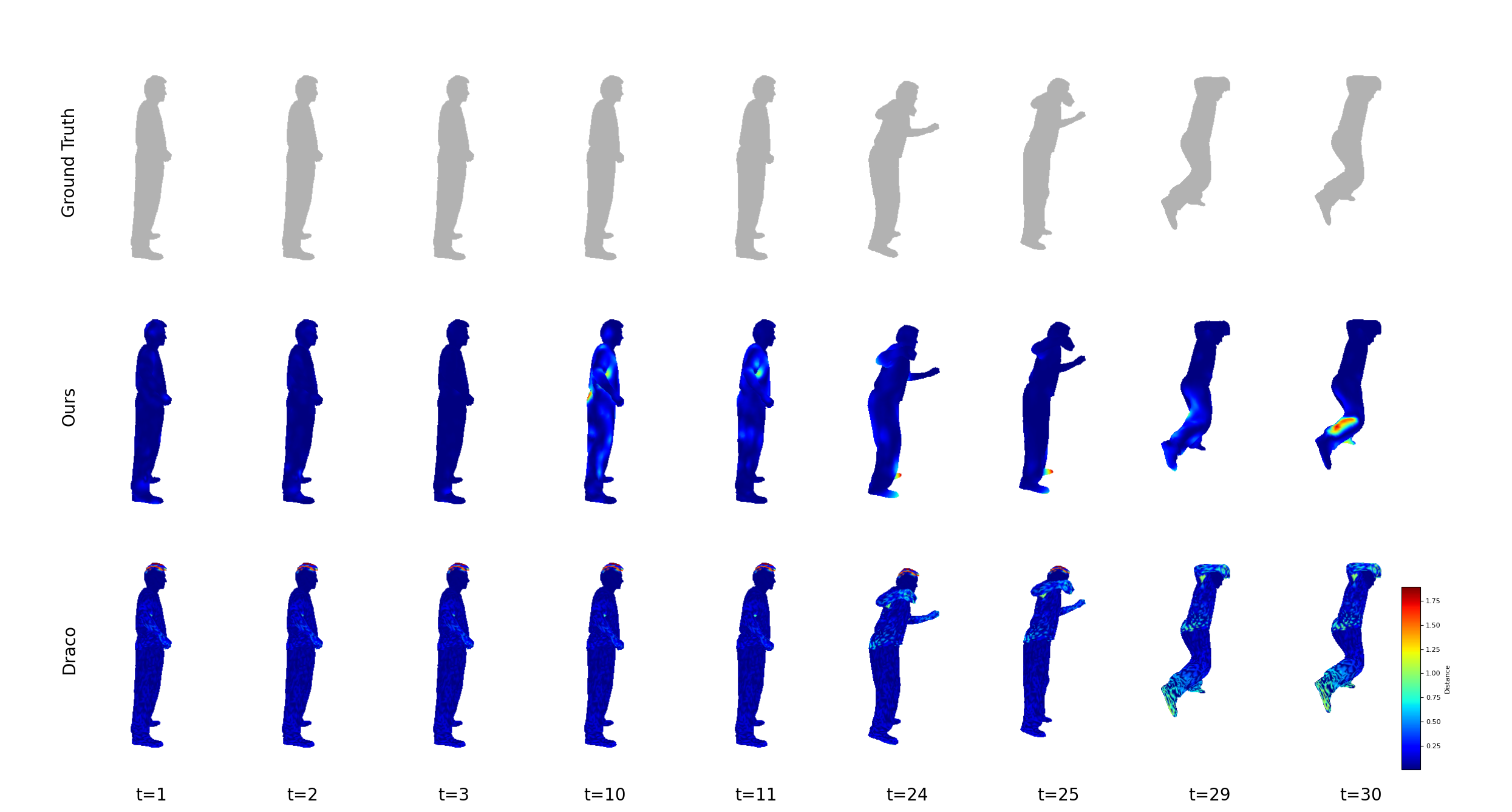}
    \caption{Visualization of Per-Vertex Distances between Ours, Draco and Groud Truth}
    \label{Visualization}
\end{figure*}

\begin{figure*}[t]

    \centering
    \begin{minipage}[b]{0.33\textwidth}
    \centering
    \includegraphics[width=1\linewidth]{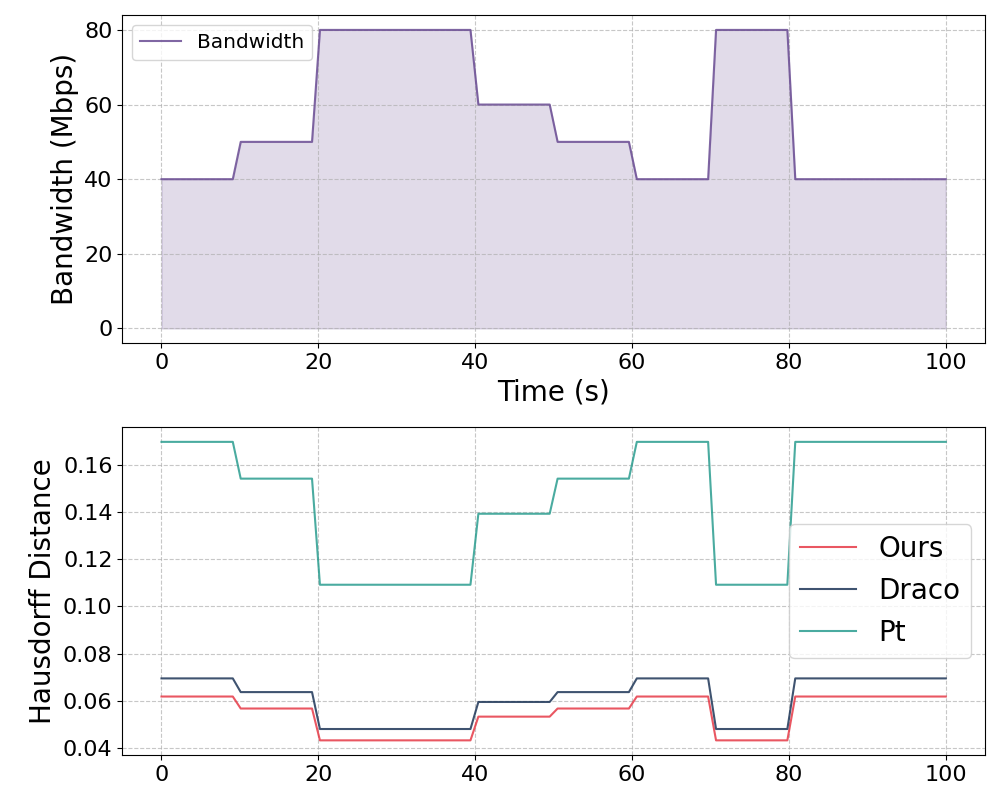}
    \caption{Bandwidth Trace and Simulation Experiment Result}
    \label{Simulation bandwidth}
    \end{minipage}
    \hfill
    \centering
    \begin{minipage}[b]{0.33\textwidth}
    \centering
    \includegraphics[width=1\linewidth]{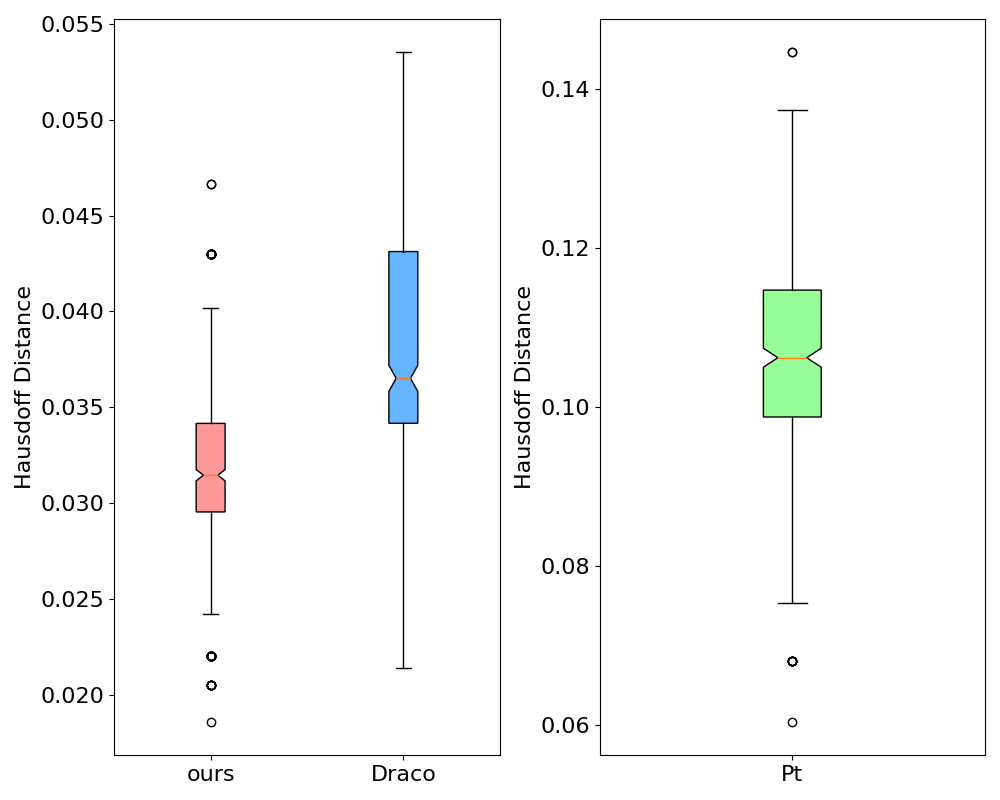}
    \caption{Simulation Experiment Result Error Distribution}
    \label{Simulation Error Distribution}
    \end{minipage}
    \hfill
    \centering
    \begin{minipage}[b]{0.33\textwidth}
    \centering
    \includegraphics[width=.8\linewidth]{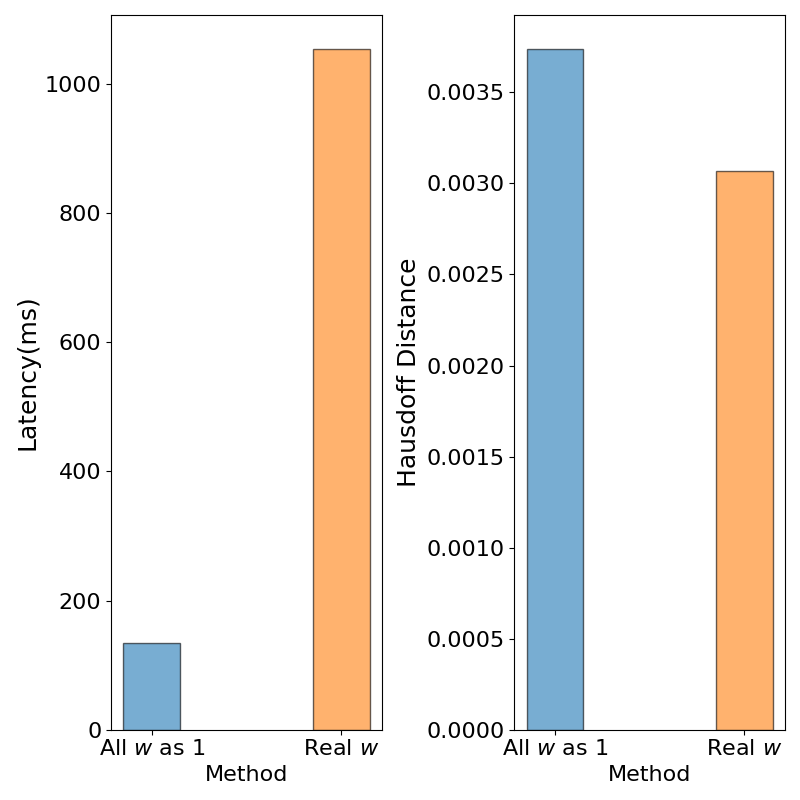}
    \caption{Ablation Study}
    \label{Ablation Study}
    \end{minipage}

\end{figure*}

\subsection{Dataset and Metrics} 
We use Peoplesnapshot~\cite{alldieck2018video} as our single-person dataset. The Peoplesnapshot dataset is a high-precision 3D human reconstruction dataset, widely used in fields such as human modeling, pose estimation, motion capture, and virtual reality. It provides highly detailed geometric data and high-quality texture information, ensuring that the 3D models are both geometrically and visually realistic. 

\head{Hausdoff Error} We use Hausdorff Error to evaluate the accuracy of a reconstructed or approximated mesh compared to a reference mesh. It measures the maximum distance between two sets of points, typically representing the vertices of the meshes in question. Specifically, it quantifies the largest distance from any point on one mesh to the closest point on the other mesh, providing a worst-case scenario metric for the mesh approximation error. The formula to calculate error between two meshes \( M_1 \) and \( M_2 \) is defined as:
\[
H(M_1, M_2) = \max \left( \max_{x \in M_1} \min_{y \in M_2} \|x - y\|, \max_{y \in M_2} \min_{x \in M_1} \|y - x\| \right)
\]
where \( \|x - y\| \) represents the Euclidean distance between points \( x \) and \( y \).


\subsection{Baseline algorithms}
\textbf{Draco:} An open-source library developed by Google for efficiently compressing and decompressing 3D geometric data, particularly for 3D meshes and point clouds. It significantly reduces file sizes while preserving data quality, making it ideal for applications like WebGL, virtual reality, and augmented reality that require fast loading times. By employing geometry and attribute compression techniques, Draco optimizes 3D data storage and transmission, making it highly useful in scenarios with limited bandwidth or storage capacity.

\textbf{Point-Cloud based Method:} We also construct a live 3D reconstruction system by point cloud. It is modified from LiveScan3D~\cite{kowalski_livescan3d_2015}, but we focus more on the compression and video streaming aspects and ultimately use Draco for encoding and compressing the point cloud.

\subsection{Qualitative Experiment}
In this section, we provide a visual impressions of compression quality by our compression method in each Frame. 

In Figure \ref{Visualization}, our method demonstrates strong performancein  compression tasks, consistently achieving lower geometry errors than Draco across most instances. This indicates that our approach is not only efficient in compressing mesh data but also effective in preserving the accuracy of the reconstructed models. In contrast, Draco, while widely used, tends to exhibit visible distortions, particularly in areas that require higher fidelity such as the head and joints. These distortions are especially pronounced during dynamic movements, where the accuracy of the mesh is critical for capturing the finer details of the animation. The results highlight that incorporating both node-based compression and transformation techniques in our method leads to a more precise mesh reconstruction compared to Draco, which relies primarily on geometric compression. This advantage is particularly evident in scenarios involving complex articulations, where maintaining the structural integrity of the mesh is paramount.

\subsection{Quantitative Experiment}
In this section, we evaluate the quantifying data by three compression methods. 

We first test our compression algorithm, Draco, and point-cloud-based method at chunk level with 30 FPS, and plotted Rate-Distortion (R-D) curves to assess the performance across various bitrates in Figure~\ref{Rate Distortion Curve}. The R-D curve illustrates the relationship between the compression rate and the Hausdorff Distance.

R-D curves indicate that while all three algorithms experience increased distortion as compression rates rise, our algorithm control distortion more effectively, particularly at higher compression rates, where the asymptote of our R-D curve is lower. This suggests that our method can maintain lower distortion when achieving higher compression.

To further quantify the efficiency of these algorithms, we calculated the Bjøntegaard Delta Rate (BD-Rate), which measures the bitrate savings of an algorithm compared to another at the same level of distortion. The BD-Rate analysis shows that our method achieves a 5.59\% bitrate savings compared to Draco and a 47.35\% savings over the point-cloud-based approach. This means that our algorithm significantly reduces the required bitrate at equivalent distortion levels, especially when compared to the point cloud method, where the savings are particularly pronounced.

We then analyze the Latency versus different sizes with our compression algorithm, Draco, and point-cloud-based method. From \ref{Latency Performance} shows that while our algorithm takes slightly longer, it remains under 600ms, which is close to real-time processing requirements. This indicates that our method provides comparable performance to Draco while delivering better compression results, almost achieving real-time performance. Additionally, compared to the point-cloud-based approach, our latency is slightly higher, but our algorithm demonstrates a clear advantage in terms of overall compression efficiency and quality.

\subsection{Simulation Experiment}
We also conducted evaluations under a simulated environment to assess the performance of the compression methods in fluctuating network conditions. The bandwidth dataset for this simulation is based on a 4G/LTE network model \cite{vanderHooft2016}. Figures \ref{Simulation bandwidth} and \ref{Simulation Error Distribution} illustrate the performance of three compression methods—ours, Draco, and a Point-cloud-based approach—under variable bandwidth conditions. Reconstruction accuracy is evaluated using the Hausdorff distance. The objective of this experiment is to analyze how bandwidth fluctuations affect the performance of each method while preserving geometric fidelity.

In Figure \ref{Simulation bandwidth}, we examine the Hausdorff distance over time for the three methods. The top subplot visualizes the bandwidth variations, ranging from 40 Mbps to 80 Mbps, which explain the observed performance fluctuations. In the bottom subplot, both our compression method and Draco demonstrate the ability to maintain relatively low and stable Hausdorff distances despite network fluctuations. In contrast, the Point-cloud-based method is more affected by bandwidth variability. The Hausdorff distances for both our method and Draco remain below 0.07, indicating that these methods are robust enough to deliver accurate volumetric video based on mesh data, even when bandwidth is inconsistent. However, the Point-cloud-based method is significantly more sensitive to network fluctuations, struggling to maintain accuracy during periods of lower bandwidth, reflecting its dependence on higher network resources for consistent performance.

Additionally, we compared the statistical distribution of Hausdorff distances for each method under different network conditions, as shown in Figure \ref{Simulation Error Distribution}. Our method exhibits the most concentrated distribution, with a median close to 0.03, indicating that it consistently achieves low Hausdorff distances across varying time intervals. In contrast, Draco shows a wider distribution and a slightly higher median, though its performance remains comparable to ours. The Point-cloud-based method, however, exhibits a much broader distribution and a higher median, highlighting its instability and greater sensitivity to network fluctuations. This demonstrates that our method is more stable and reliable in maintaining geometric accuracy, especially under challenging network conditions.

\subsection{Ablation Study}
Figure \ref{Ablation Study} presents a performance comparison of our method under two distinct scenarios: one in which all weights $w$ are uniformly set to 1, and another where the actual, computed weights are preserved. Both evaluations were conducted under consistent conditions, with a fixed bandwidth of 100 Mbps and a chunk size of 30. The focus of the comparison lies on two critical metrics: latency, represented by the total processing time, and geometric accuracy, measured using the Hausdorff distance.

In terms of latency, the approach where all weights $w$ are set to 1 demonstrates a substantial performance improvement, achieving a total processing time of just 0.13 seconds, nearly reaching real-time performance. In contrast, using the actual weights results in a significantly longer processing time of 2.1 seconds. This substantial reduction in latency when simplifying the weights underscores the potential of the uniform-weight approach for applications where speed is of paramount importance.

Despite this improvement in speed, the trade-off in geometric accuracy is minimal. While the method using uniform weights does result in a slightly higher Hausdorff distance, indicating a marginal increase in geometric error, the difference is not significant enough to affect the overall visual quality in most practical scenarios. The model retains a high level of fidelity, particularly in the context of video playback or real-time rendering.

The results indicate that setting all $w$ to 1 offers a viable balance between computational efficiency and accuracy, making it an ideal choice for scenarios that prioritize speed while still maintaining a satisfactory level of visual quality.

%% file: 6-related.tex
\section{Related Works}
\head{Volumetric Video Streaming Systems}
Recent research has explored various methods for point-cloud-based volumetric video streaming. ViVo~\cite{han_vivo_2020} employs a FoV-based approach to minimize the number of points transmitted to viewers but does not account for real-time capture scenarios. YuZu~\cite{zhang_yuzu_nodate} uses a neural network to compress point clouds for streaming, however, it necessitates pre-training a model for each video, requiring hours of training. CaV3~\cite{liu_cav3_2023} leverages buffering and field-of-view (FoV) prediction to optimize the streaming process. Holoportation~\cite{orts-escolano_holoportation_2016} utilizes a lightweight compression to compress texture and geometry data to enable real-time scene capture, though it demands 1-2 Gbps bandwidth per scene. FarfetchFusion~\cite{lee_farfetchfusion_2023} uses a temporality similarity-based method to stream the 3D face, but due to the limitation of their facial landmark detection mechanism, it is naturally only applicable to the scenario of face-to-face telepresence. While these works have advanced volumetric video streaming, challenges remain in mesh-representation-based streaming systems.

\head{Time-Varing Mesh Compression}
Existing time-varying mesh compression codecs can be categorized based on their approach to handling fixed or varying connectivity, such as prediction-based methods~\cite{hajizadeh_nlme_2020}, PCA~\cite{lalos_feature_2018,lalos_adaptive_2017}, segmentation~\cite{luo_3d_2019,luo_spatio-temporal_2020}, and wavelet transforms~\cite{payan_temporal_2007,boulfani-cuisinaud_motion-based_2007}. While these techniques are effective in minimizing per-frame data, they do not address the specific challenges associated with streaming dynamic meshes in a network-adaptive manner. In streaming scenarios, it's crucial to consider network conditions and playback buffer metrics (e.g. encoding and decoding time), which these methods often overlook.

\head{Deformation Models and Registration}
Non-rigid registration through methods like ICP~\cite{besl_method_1992,amberg_optimal_2007}, has been a foundation for deformation models, focusing on local proximity and general deformations. RPM~\cite{gold_new_1998} improves on ICP by avoiding local minima, similar to our optimization of visual quality and bandwidth through our QoE model. Their focus is to registrate one surface to another. Our work extends deformation models, such as those assuming isometry~\cite{chen_efficient_2023} or using Gaussian mixtures~\cite{jian_robust_2011}, by leveraging embedded deformation to control frame-to-frame coherence with fewer nodes, balancing computational efficiency and bandwidth usage. Unlike cage-based approaches~\cite{hajizadeh_nlme_2020}, which struggle with subtle changes, we dynamically adjust mesh nodes to enhance quality. We also build on deformation graph methods~\cite{sumner_embedded_2007}, where affine transformations ensure smooth deformations, and optimized for real-time mesh streaming under fluctuating network conditions.

%% file: 7-conclusion.tex
\section{Conclusion}
In conclusion, \sysname presents a transformative approach to volumetric video streaming by leveraging mesh-based data deformability to enhance both bandwidth efficiency and visual quality. By embedding deformation within the streaming pipeline, the framework addresses critical challenges in real-time transmission, such as excessive bandwidth consumption and latency, which are prevalent in traditional streaming methods. The innovative QoE model and dynamic programming algorithm introduced by \sysname ensure adaptability to varying network conditions, offering a robust solution that outperforms existing systems.